\documentclass[10pt,twocolumn,letterpaper]{article}

\makeatletter
\@namedef{ver@everyshi.sty}{}
\makeatother
\usepackage{cvpr}              

\usepackage{graphicx}
\usepackage{amsmath}
\usepackage{amssymb}
\usepackage{booktabs}

\usepackage[pagebackref,breaklinks,colorlinks]{hyperref}

\usepackage{wrapfig}
\usepackage{enumitem}
\usepackage{subcaption}
\usepackage{booktabs}
\usepackage{multirow}
\usepackage[accsupp]{axessibility}
\usepackage{array}
\usepackage[dvipsnames,table,xcdraw]{xcolor}
\usepackage{graphicx}
\usepackage{tikz}
\usetikzlibrary{shapes}

\newcommand{\tableCellHeight}{1}
\newcommand{\tabstyle}[1]{
  \setlength{\tabcolsep}{#1}
  \renewcommand{\arraystretch}{\tableCellHeight}
  \centering
  \small
}

\setlist[itemize]{leftmargin=*}

\newcommand{\hgreen}[1]{\textcolor{ForestGreen}{\textbf{#1}}} 
\definecolor{tabhighlight}{HTML}{e5e5e5}
\definecolor{citecolor}{HTML}{0071bc}

\newcommand{\shortName}{LASP}
\newcommand{\shortNameZero}{LASP-V}

\definecolor{darkcandyapplered}{rgb}{0.64, 0.0, 0.0}
\definecolor{burgundy}{rgb}{0.5, 0.0, 0.13}
\definecolor{carnelian}{rgb}{0.7, 0.11, 0.11}

\usepackage[capitalize]{cleveref}
\crefname{section}{Sec.}{Secs.}
\Crefname{section}{Section}{Sections}
\Crefname{table}{Table}{Tables}
\crefname{table}{Tab.}{Tabs.}

\def\cvprPaperID{3856} 
\def\confName{CVPR}
\def\confYear{2023}

\newcommand*{\affaddr}[1]{#1} 
\newcommand*{\affmark}[1][*]{\textsuperscript{#1}}

\begin{document}

\title{LASP: Text-to-Text Optimization for Language-Aware Soft Prompting \\ of Vision \& Language Models}

\author{%
Adrian Bulat\affmark[1,2], Georgios Tzimiropoulos\affmark[1,3]\\
\affaddr{\affmark[1]Samsung AI Cambridge}\;\;
\affaddr{\affmark[2]Technical University of Iasi}\;\;
\affaddr{\affmark[3]Queen Mary University of London}
}

\maketitle

\begin{abstract}
Soft prompt learning has recently emerged as one of the methods of choice for adapting V\&L models to a downstream task using a few training examples. However, current methods significantly overfit the training data, suffering from large accuracy degradation when tested on unseen classes from the same domain.  To this end, in this paper, we make the following 4 contributions: (1) To alleviate base class overfitting, we propose a novel Language-Aware Soft Prompting (LASP) learning method by means of a text-to-text cross-entropy loss that maximizes the probability of the learned prompts to be correctly classified with respect to pre-defined hand-crafted textual prompts. (2) To increase the representation capacity of the prompts, we propose \textit{grouped} LASP where each group of prompts is optimized with respect to a separate subset of textual prompts. (3) We identify a visual-language misalignment introduced by prompt learning and LASP, and more importantly, propose a re-calibration mechanism to address it. (4) We show that LASP is inherently amenable to including, during training, \textit{virtual classes}, i.e. class names for which no visual samples are available, further increasing the robustness of the learned prompts. Through evaluations on 11 datasets, we show that our approach (a) significantly outperforms all prior works on soft prompting, and (b) matches and surpasses, for the first time, the accuracy on novel classes obtained by hand-crafted prompts and CLIP for 8 out of 11 test datasets. Code will be made available \href{https://www.adrianbulat.com/lasp}{here}.
\end{abstract}

\section{Introduction}

Large-scale pre-training of neural networks has recently resulted in the construction of a multitude of foundation models for Language~\cite{devlin2018bert,radford2019language} and Vision \& Language (V\&L) understanding~\cite{radford2021learning,jia2021scaling,yu2022coca, alayrac2022flamingo}. Unlike the previous generation of neural networks, such models can better capture the distribution of the world from which new favorable properties and characteristics emerge. Of particular interest to this work are V\&L models trained with contrastive learning (i.e. CLIP-like models~\cite{radford2021learning,jia2021scaling,li2021supervision,yao2021filip,yu2022coca}), which have enabled seamless few-shot and even zero-shot adaptation to new downstream tasks and datasets. Specifically, this paper proposes a simple yet highly effective way to drastically improve soft prompt learning for the few-shot adaptation of the V\&L model to a given downstream task.

Similarly to their NLP counterparts~\cite{radford2021learning,lester2021power,li2021prefix}, prompt engineering and learning has emerged as one of the most powerful techniques for adapting a V\&L to new tasks. Initially, in~\cite{radford2021learning}, a set of manually-defined hand-engineered templates (or prompts) like \texttt{a photo of a \{cls\_name\}}, or \texttt{a black and white photo of a \{cls\_name\}} were passed through the text encoder of the V\&L model to create class-specific weights for category \texttt{cls\_name} that can be used for zero-shot recognition. Following research in NLP~\cite{lester2021power,li2021prefix}, subsequent work~\cite{zhou2022learning,zhou2022conditional} has proposed replacing the manually picked templates with a sequence of learnable vectors, also coined \textit{soft prompts}, which are fed as input to the text encoder along with the class name \texttt{cls\_name}. The soft prompts are learned from a few training examples with the entire V\&L model kept frozen. The whole process can be seen as parameter efficient fine-tuning of the model on a small training dataset. 

However, a clearly identifiable problem with prompt learning is base class overfitting: while the accuracy on the classes used for training (base classes) significantly increases, the accuracy on unseen, during training, (novel) classes significantly drops. This is to some extent expected, as soft prompts are learned from few examples belonging to the base classes. Notably, on novel classes, direct, zero-shot recognition using hand-engineered prompts outperforms all existing soft prompt learning methods.

\noindent \textbf{Key idea:} To alleviate base class overfitting, in this work, we propose a solution motivated by the following observation: since prompt learning improves the accuracy on base classes, but prompt engineering is significantly better on novel classes, we propose to learn the soft prompts by adding a cross entropy text-to-text loss that enforces the learned prompts to be close, in embedding space, to the textual ones, thus exploiting the intrinsic information captured by the text encoder. The proposed text-to-text loss enables language-only optimization for V\&L model adaption for the first time. This is in contrast with prior soft-prompt learning methods that only capture V\&L interactions. 

\noindent \textbf{Key contributions:} Based on the above, we propose a novel framework for soft prompt learning which we call Language-Aware Soft Prompting (LASP). Our main contributions within the LASP framework are as follows:
\vspace{-0.15cm}
\begin{itemize}[itemsep=-0.3em]
    \item We propose, for the first time, language-only optimization for V\&L model adaption. Specifically, we propose a novel text-to-text cross-entropy loss that maximizes the probability of the learned prompts to be correctly classified with respect to the hand-engineered ones and show its effectiveness in terms of alleviating base-class overfitting. 
    \item
    To increase the representation capacity of the prompts, and inspired by grouped convolution and multi-head attention, we propose a grouped language-aware prompt representation where \textit{each group} of prompts specializes to a different subset of the pre-defined manual templates.
    \item 
    We identify a visual-language misalignment introduced by prompt learning and LASP which impacts the generalization. More importantly, we propose a re-calibration mechanism based on (a) Layer Normalization fine-tuning and (b) learning a class-agnostic bias to address it. 
    \item 
    Thanks to our language-only learning framework, we propose training LASP with virtual classes by including, during training, class names for which no visual samples are available. Importantly, we show that this further increases the robustness of the learned prompts. 
\end{itemize}

\noindent \textbf{Main results:} Our methods set a new state-of-the-art for few-shot and zero-shot image classification on 11 datasets, significantly outperforming all soft prompting prior works. Importantly, we present, for the first time, a prompt learning method that outperforms, for the majority of the test datasets (8 out of 11), the very strong baseline based on hand-crafted prompts and CLIP for the recognition of novel classes (i.e. zero-shot setting).

\section{Related work}

\noindent \textbf{Contrastive V\&L Models:} Recently, large scale V\&L pre-training with contrastive learning has been used to train foundation models resulting in robust representations, transferable to new tasks both under few-shot and zero-shot  settings~\cite{radford2021learning,jia2021scaling,li2021supervision,yao2021filip,yu2022coca}. Such networks consist of a vision encoder (typically a ViT~\cite{dosovitskiy2020image}) and a Transformer-based text encoder~\cite{vaswani2017attention}. Highly parameterized instantiations of such architectures are trained on large corpora of image-caption pairs (e.g.~\cite{radford2021learning} uses 400M and~\cite{jia2021scaling} 1B pairs) using contrastive learning. We used CLIP~\cite{radford2021learning} as the foundation model for our method. 

\noindent \textbf{Prompt Learning} is about adapting pre-trained foundational models on (downstream) tasks, typically in a zero-shot or few-shot setting. Firstly proposed in the context of Language Models (LM), prompting was initially about prepending hand-crafted instructions/examples to the task input so that the LM generates the appropriate output conditioned to the input~\cite{radford2019language,brown2020language}. In~\cite{schick2020exploiting, schick2020s}, the main idea is to reformulate the downstream task as a \textit{cloze} task using hand-crafted patterns (or templates), thus avoiding the need to train a task-specific classifier. As finding the optimal patterns is laborious, recent works have attempted to address this by learning a set of soft (continuous) prompts~\cite{lester2021power,li2021prefix}. 

In V\&L foundation models, like CLIP, the class names are used to create hand-crafted prompts~\cite{radford2021learning} that are fed as input to the text encoder, enabling zero-shot visual recognition. CoOp~\cite{zhou2022learning} extends work on soft prompt optimization to the V\&L domain by learning a set of $M$ prompts which are used as input to the text encoder alongside the class name. The prompts are learned by minimizing the classification error on a training set consisted of the given base classes. One major limitation of CoOp is weak generalization: the learned prompts overfit the base classes and do not work well when tested on novel classes. To alleviate this, CoCoOp~\cite{zhou2022conditional} proposes a dynamic version of~\cite{zhou2022learning} where a small network is trained to produce a visual feature from the input image that is added to the learned prompts, hence making them input specific (i.e. dynamic). 
ProDA~\cite{lu2022prompt} adopts a probabilistic approach 
by modelling the distribution of the prompts at the output of the text encoder as a multivariate Gaussian distribution. The estimated mean is used during inference. Finally, UPL~\cite{unsup_prompt22} uses CLIP to generate pseudo-labels on the target dataset and then self-training to learn the soft prompts. 
Finally, ProGrad~\cite{zhu2022prompt} aims to adapt the V\&L model to each target domain by encouraging it ``not to forget'' CLIP’s zero-shot predictions using a KL visual-text loss between the CLIP’s logits and their model’s logits (\ie they use visual features). The weights are then updated in the direction perpendicular to CLIP gradients. In contrast, our loss is
a pure text-to-text loss, further allowing for the incorporation of virtual classes. Unlike~\cite{zhu2022prompt}, we outperform CLIP on novel classes.

The proposed LASP framework alleviates base class overfitting and significantly improves upon the previously reported best results without resorting to a dynamic approach as in CoCoOp~\cite{zhou2022conditional}. In its basic version, LASP deploys a text-to-text loss that enforces the learned prompts to be ``close'' to a set of manually defined textual prompts in the text encoder space. Importantly, the basic LASP can be extended in three important ways: (1) by allowing the incorporation of virtual classes, i.e. novel class name information for which no (visual) training data is available (LASP-V). This is shown to significantly improve the robustness of the learned prompts at no extra cost during inference; (2) by allowing the use of a grouped prompt representation within the proposed language-aware training which is shown to increase the representation capacity of the learned prompts; (3) by performing further optimization of the visual encoder so that the visual and text embeddings are realigned resulting in significant accuracy gains. Notably, our approach is very efficient (as efficient as~\cite{zhou2022learning}) as opposed to~\cite{zhou2022conditional} which requires recomputing all the class-related text embeddings every time a new image is to be classified.  

\begin{figure*}[!ht]
    \centering
    \includegraphics[trim={0cm 7.0cm 4cm 0},clip,width=0.88\textwidth]{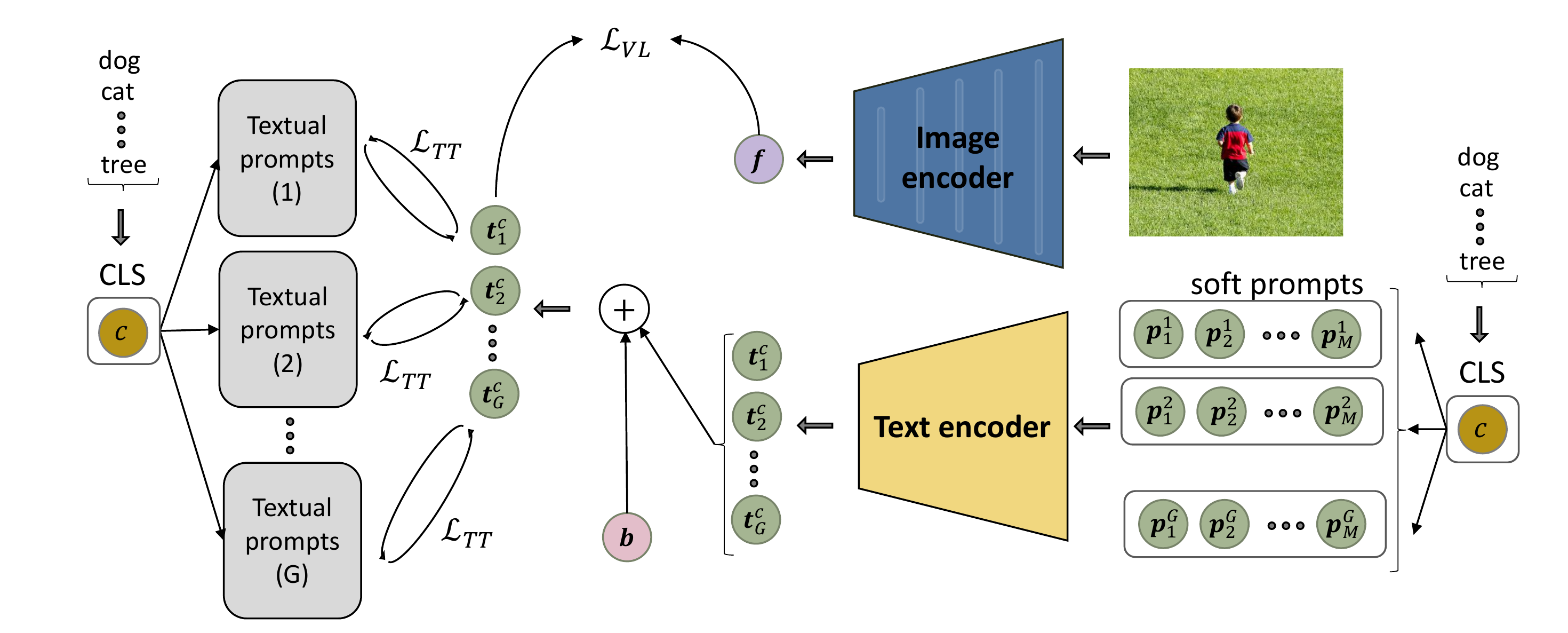}
    \caption{\textbf{Overall idea.} 
    While standard prompt learning is based on image-text interactions ($L_{VL}$ loss; Eq.~\ref{eq:Lvl}), LASP additionally models text-text interactions using the proposed Text-to-Text loss $L_{TT}$ (Eq.~\ref{eq:Ltt}). There are $G$ groups of learned prompts $\mathbf{p}_i^j$ passed through the text encoder to form $G$ text embeddings $\mathbf{t}_j$ summarizing the input. The $L_{TT}$ loss is then applied over the different groups of the text embeddings and the textual prompts. Moreover, to alleviate data distribution shift and visual-language misalignment, the LN layers of the visual encoder are fine-tuned and the embeddings are ``corrected'' at the output space by the learnable vector $\mathbf{b}$, shared for all classes. The text encoder remains entirely frozen. Notably, LASP can be trained with virtual classes by including, during training, class names for which no visual samples are available.} 
    \label{fig:main_idea}
    \vspace{-5mm}
\end{figure*}

\section{Method}

\subsection{Background} \label{ssec:prelim}

\noindent \textbf{Prompt engineering} enables zero-shot visual recognition using V\&L models trained with contrastive learning (CLIP in this work) as follows: Given a set $\mathcal{V}$ of $C$ class names, $\texttt{class\_name}_c$, $c \in \{1,\dots, C\}$, a prompt, i.e. a manually designed template concatenated with the class name like $h_c=$\texttt{a photo of a $\{\texttt{class\_name}_c\}$}, is passed  
through the V\&L's text encoder $g_T(.)$ to compute the class specific text feature (weight) $\mathbf{t}^h_c = g_T(h_c)$. Moreover, an image $\textbf{x}$ to be classified is passed through the V\&L's image encoder $g_I(.)$ to compute image specific feature $\textbf{f} = g_I(\mathbf{x})$. A probability distribution over the class labels is given by:
\begin{equation}
    P_h(y|\mathbf{x}) =  \frac{\exp \Bigl( \texttt{cos}(\mathbf{t}^h_y, \mathbf{f}) / \tau \Bigr) }{\sum_{c=1}^{C} \exp \Bigl(\texttt{cos}(\mathbf{t}^h_c, \mathbf{f})/\tau\Bigr)},
\end{equation}
where $\tau$ is a temperature factor and $\texttt{cos}$ the cosine similarity. Finally, the class for $\textbf{x}$ is given by $\tilde{y}=\arg_{max} P_h(y|\mathbf{x})$. Note that, to compute $\mathbf{t}^h_c$, no training with class specific image data is required, thus enabling zero-shot recognition for any given class name.

\noindent \textbf{Soft prompt learning}~\cite{lester2021power,li2021prefix, zhou2022learning} is concerned with parameter efficient fine-tuning of a pre-trained V\&L model by learning a sequence of $M$ learnable vectors $\mathbf{p}_m\in \mathbb{R}^{d}, m=\{1,\dots,M\}$ using a few labelled samples. Specifically, the manually picked prompt $h_c$ is replaced by a new learnable one $\mathbf{r}_c$ formed by concatenating the sequence of $\mathbf{p_m}$ with the word embedding $\textbf{w}_c$ of $\texttt{class\_name}_c$, that is: $\mathbf{r}_c = \{\textbf{p}_1, \textbf{p}_2, \dots, \textbf{p}_M, \textbf{w}_c\}$, and, finally, a class specific text feature  $\mathbf{t}^r_c = g_T(\mathbf{r}_c)$ is obtained. A probability distribution over the class labels is:
\begin{equation}
    P_r(y|\mathbf{x}) =  \frac{\exp \Bigl( \texttt{cos}(\mathbf{t}^r_y, \mathbf{f}) / \tau \Bigr) }{\sum_{c=1}^{C} \exp \Bigl(\texttt{cos}(\mathbf{t}^r_c, \mathbf{f})/\tau\Bigr)}.
\end{equation} 
The prompts can be learned by minimizing the cross-entropy loss:
\begin{equation}
    \mathcal{L}_{VL} = - \sum_{c=1}^C \log P_r(c|\mathbf{x}) y_{c}. \label{eq:Lvl}
\end{equation}
Note that the V\&L model remains entirely frozen during training. Moreover, as the soft prompts are typically shared across all classes, they can be directly used for zero-shot evaluation on additional novel classes.

\subsection{Language-Aware Soft Prompting (LASP)}\label{ssec:mehod-lasp}

Despite its strong performance on base classes, vanilla soft prompt learning (see Sec.~\ref{ssec:prelim}) under-performs on novel classes (i.e. zero-shot setting). While CoCoOp~\cite{zhou2022learning} partially alleviates this by conditioning on the image feature, its accuracy for the zero-shot setting is still trailing that of CLIP with hand-crafted prompts. Moreover, it requires passing the prompts for all classes through the text encoder every time a new image is to be classified. 

In this work, we propose, for the first time, language-only optimization for V\&L downstream adaption. This is in contrast with prior soft-prompt learning methods that only capture V\&L interactions. Specifically, since the hand-engineered textual prompts outperform the learnable soft prompts for the zero-shot setting, then, in order to avoid base-class overfitting and strengthen generalizability, we propose that the learnable ones should be trained so that they can be correctly classified in language space where the class weights are given by the textual prompts. In other words, the model is forced to correctly classify the learnable prompts into the corresponding hand-crafted ones.

To this end, a second cross entropy loss is used to minimize the distance between the encoded learned soft prompts and the encoded textual ones. Specifically, recall that $\mathbf{t}^h_c = g_T(h_c)$ is the class weight for class $c$ obtained by encoding the corresponding textual prompt. Assuming that $L$ manually defined textual prompts are available~\footnote{The original CLIP prompts serve as textual prompts without any tweaking or change. Note, that our method can even work with random sentences (see Sec.~\ref{ssec:ablation}).}, we have $\mathbf{t}^{h,l}_c, \;l=1,\dots,L.$ Moreover, $\mathbf{t}^r$ is an encoded learnable prompt to be classified in one of the $C$ classes. Finally, the probability of prompt $\mathbf{t}^r$ being classified as class $y$ is:
\begin{equation}
    \label{eq:p_t}
    P_{rh}(y|\mathbf{t^r}) = \frac{1}{L}\sum_{l=1}^L \frac{\exp \Bigl( \texttt{cos}(\mathbf{t}_{y}^{h,l}, \mathbf{t}^r) / \tau \Bigr) }{\sum_{c=1}^{C} \exp \Bigl(\texttt{cos}(\mathbf{t}_{c}^{h,l}, \mathbf{t}^r)/\tau\Bigr)}.
\end{equation}
The language-aware training loss is computed similarly to the V\&L loss:
\begin{equation}
    \mathcal{L}_{TT} = - \sum_{c=1}^C \log P_{rh}(c|\mathbf{t}^r) y_{c}, \label{eq:Ltt}
\end{equation}
with the overall training objective defined as:
\begin{equation}
    \mathcal{L} =  \alpha_{VL} \mathcal{L}_{VL} + \alpha_{TT} \mathcal{L}_{TT},
    \label{eq:loss_combined}
\end{equation}
where $\alpha_{VL}$ and $\alpha_{TT}$ are user-defined scaling coefficients controlling the magnitude of the $\mathcal{L}_{VL}$ and $\mathcal{L}_{TT}$ losses, respectively. Overall, we call the proposed learning formulation Language-Aware Soft Prompting (\shortName). See also Fig~\ref{fig:main_idea}.
\textbf{Interpretations:} LASP can be interpreted in a number of ways:

\noindent \textit{\textbf{\shortName~as a regularizer:}} Although the learned prompts constitute a small number of parameters, especially in the few-shot setting, the resulting models (prompts) are prone to overfitting to base classes~\cite{zhou2022learning}. 
As the proposed language-aware loss encourages the learned prompts to be close in embedding space to the textual ones, LASP can be naturally viewed as a regularizer that prevents the learned prompt-conditioned features from diverging too much from the hand-crafted ones.

\noindent \textit{\textbf{\shortName~as language-based augmentation:}} Current soft prompt learning methods restrict augmentation to the vision domain, where random transformations, such as rotation, color jittering or scaling, increase the robustness of the system, especially for cases with limited number of training samples. However, no augmentations are performed in the language domain. Ideally, we want the prompt-conditioned text embedding to be robust too, capturing the full space of each class. In practice, we can achieve this by targeted prompting, where we can specify certain characteristics and/or apply text-based transformations to the class name, e.g.: ``A sketch of \textit{dog}'' or ``A rotated photo of a \textit{dog}''. 

At train time, as reflected by Eq.~\ref{eq:p_t}, we compute the class label distribution per $l-$th template and then average over all templates. Hence, we opt not to mix across templates during training as we want the model to focus on class information solely. For example, the model could distinguish easier between a ``a sketch of a \textit{dog}'' and ``a photo of a wolf'' compared to ``a sketch of a \textit{dog}'' and ``a sketch of a wolf'',  as in the former case, the style could be used as an additional queue. We validated this in preliminary experiments (intermixing the templates was found to hurt performance by 0.5\% on novel classes).

\begin{figure}[ht]
    \begin{subfigure}[b]{0.45\textwidth}
        \centering
     \includegraphics[height=2.7cm]{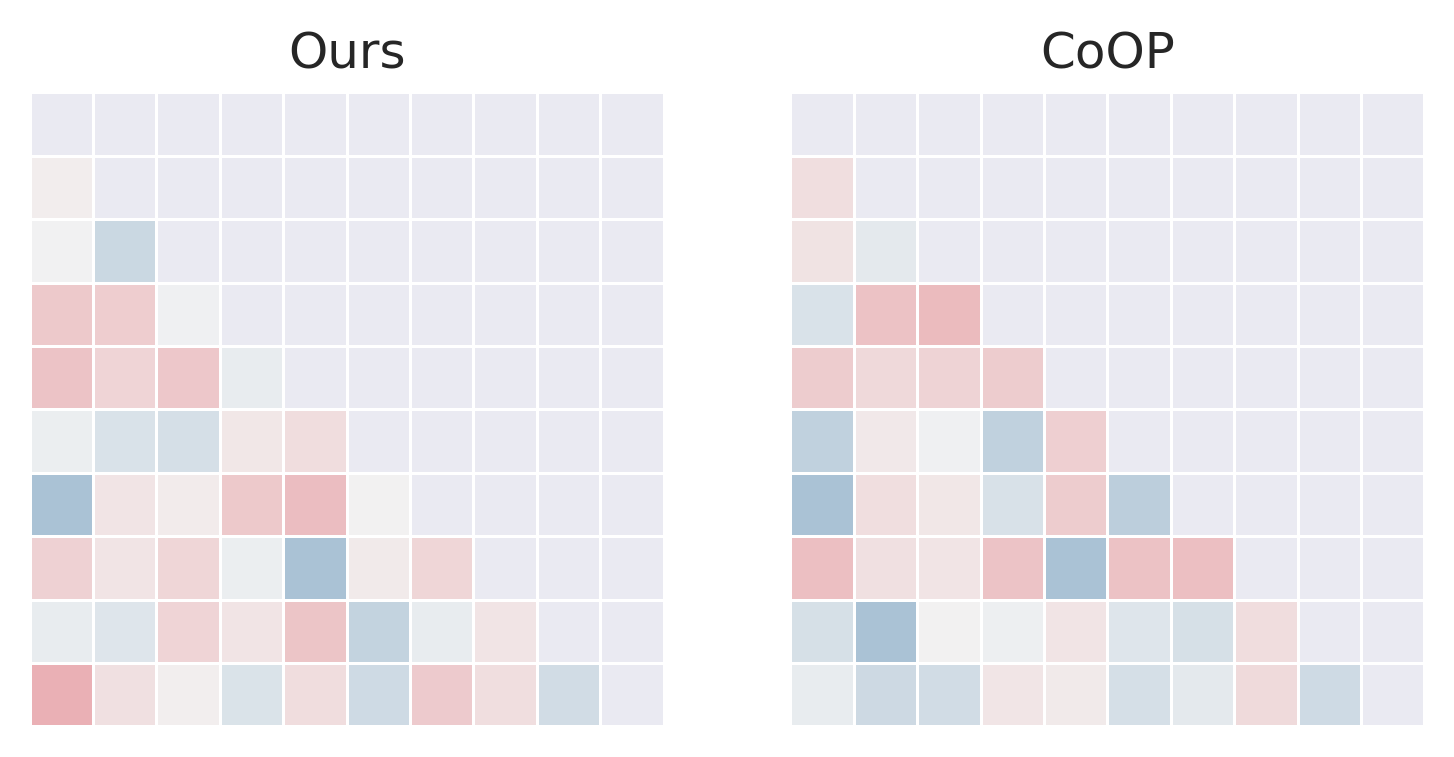}
      \caption{Eurosat; Ours ($0.516$) vs CoOp ($0.491$) }
    \end{subfigure}
    \hfill
    \begin{subfigure}[b]{0.45\textwidth}
        \centering
     \includegraphics[height=3.0cm]{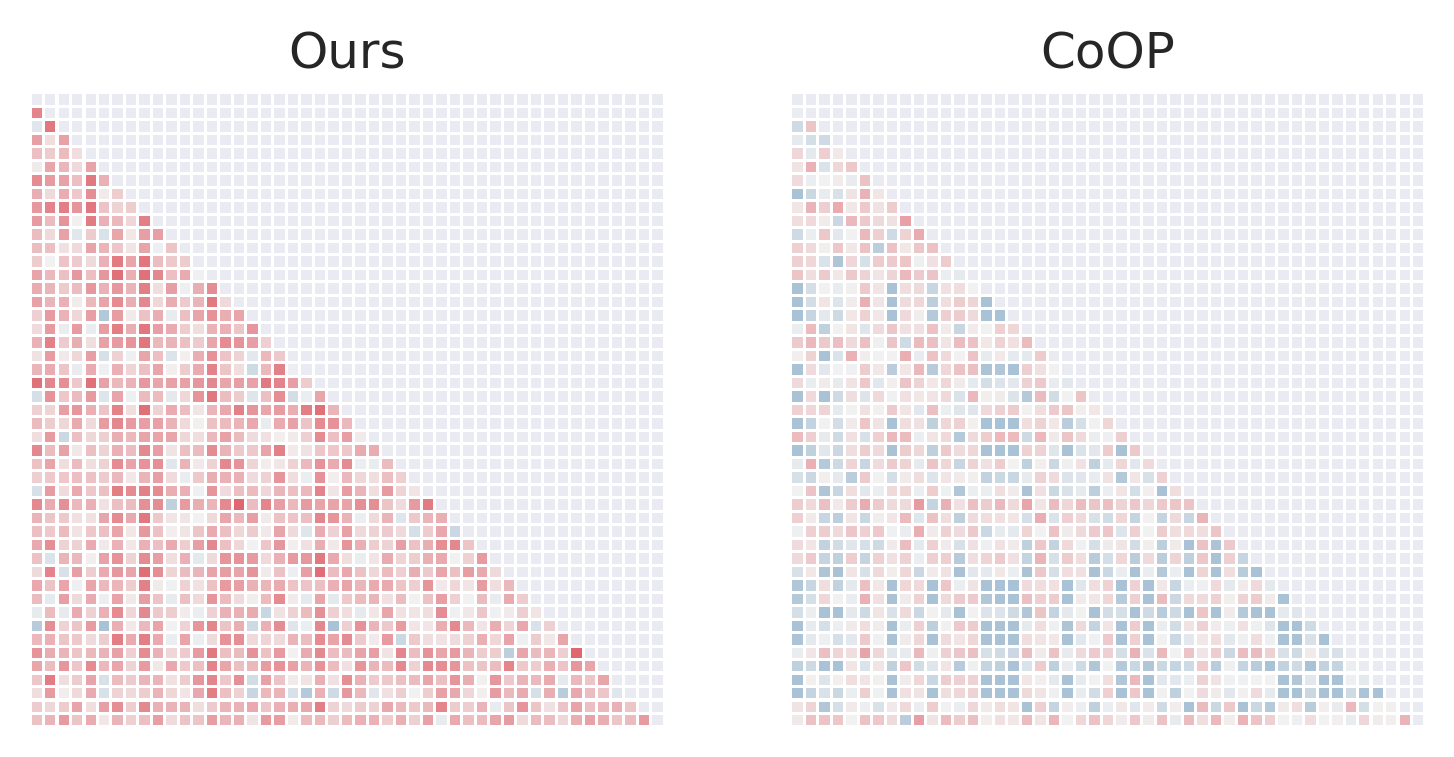}
      \caption{DTD; Ours ($0.644$) vs CoOp ($0.488$) }
    \end{subfigure}
    \caption{\textbf{Cosine distance between the class embeddings} produced by the CLIP text encoder on Eurosat and DTD for LASP and CoOp. Class centroids situated further apart are more separable as the underlying image features are identical across both methods. Brighter colors indicate bigger distances. The numbers shown are the average cosine distance between the classes. }
    \label{fig:distance_text_embeddings}
    \vspace{-0.2cm}
\end{figure}

\noindent \textit{\textbf{\shortName~for discriminative class centroids:}} By optimizing w.r.t both image and text, our method produces class centroids that are more discriminative and have a higher separation margin. This can be visualized in Fig.~\ref{fig:distance_text_embeddings} where we plot the cosine distance between the embeddings of each class. Our approach learns class centroids that have a higher cosine distance than those of our baseline, CoOp.

\noindent \textit{\textbf{LASP as data-free distillation:}} Typically, knowledge distillation requires a training set of images, where a teacher network provides a training signal for the student~\cite{hinton2015distilling}. LASP's text-to-text loss can be also interpreted as a data-free distillation (\ie does not use any image data)  where the learnable prompts define the ``samples''. As CLIP learns a joint V\&L space, similar concepts are close together across both domains. Hence, optimizing against a concept or object in the language domain, using the proposed loss, should also help make a step in the visual domain, improving the classification of the images.

\subsection{Grouped LASP}\label{ssec:method_text_grouped}

Grouped convolutions~\cite{krizhevsky2017imagenet} and multi-head attention~\cite{vaswani2017attention} have been shown to learn strong representations. The groups or the number of heads, respectively, can be also interpreted as a set of experts that are then combined to produce a strong feature. Drawing inspiration from this, we propose a grouped prompt representation, where each group is optimized with respect to a separate subset of textual prompts. Effectively, the prompts from each group will learn a transformation specialized to its corresponding subset (analogous to the aforementioned techniques that also specialize to a part of the signal). In particular, we split the set of $L$ templates into $G$ equally sized  sub-sets. Moreover, each sub-set is associated with a sequence of $M$ prompts $\mathbf{r}^g_c = \{\textbf{p}^g_1, \dots, \textbf{p}^g_M, \textbf{w}_c\}, g=1,\dots,G$ each producing a class specific text feature  $\mathbf{t}^{r,g}_c = g_T(\mathbf{r}^g_c)$. Finally, our text-to-text loss in Eq.~\ref{eq:Ltt} becomes:
\begin{equation}
    \mathcal{L}_{TT-G} = - \sum_{g=1}^G\sum_{c=1}^C \log P_{rh}^g(c|\mathbf{t}^g) y_c, \label{eq:Lttg}
\end{equation}
with $P_{rh}^g$ computed for each group similarly to Eq.~\ref{eq:p_t}. We note that the splits were created randomly. As the text templates are in general semantically independent, no preferred grouping arises.
At test time, the final result is computed by taking the average of the cosine similarity scores between each group and the visual feature $\mathbf{f}$.

\subsection{Re-aligning LASP}\label{ssec:method_vision_distribution}

\noindent \textbf{Combating data distribution shift:} for some downstream tasks, it is possible that there is a data distribution shift between the downstream image dataset and the one used by CLIP during training. Hence, we would like this aspect to be captured by the downstream adaptation method. To this end, some optimization of the visual encoder can be performed; nevertheless this can very easily result in base class overfitting if, after the training, the V\&L embeddings are pushed away from the joint space learned by CLIP. For example, preliminary results with visual adapters have shown that they hurt zero-shot accuracy. On the contrary, we found that Layer Normalization (LN)~\cite{ba2016layer} fine-tuning is a much more robust way to adapt the visual encoder. Overall, we propose fine-tuning the LN of the CLIP encoder as a way to combat distributional shift. 

\noindent \textbf{Combating V\&L misalignment:} Because after LN fine-tuning the V\&L are not guaranteed to continue to be aligned, we also propose to learn a ``correction'' at the output of the text encoder in the form of a learnable offset (bias) that aims to re-align the two modalities. Let $\mathbf{W}$ be the set of weights of the linear classifier obtained by passing the learned prompts from the text encoder. We propose to learn a vector $\mathbf{b}\in \mathbb{R}^d$ that is simply added to $\mathbf{W}$, that is $\mathbf{W} = \mathbf{W} + \mathbf{b}$. Importantly, the learned offset is shared among all classes, and in this way it can be readily applied for the case of novel classes too.

\subsection{LASP with Virtual Classes (\shortNameZero)}

A direct observation that can be drawn from Eq.~\ref{eq:p_t} is that, in practice, we do not have to use only the class names for which we have labelled image data, as the value of $L_{TT}$ is independent of the input image. To this end, we propose to learn the prompts using both annotated image-text pairs and \textit{class names} outside the base set (for which we have no images available). We call this setting as training \textit{LASP with virtual classes}. Our setting combines the best of both words: the guidance from the few annotated image samples and the zero-shot generalizability of language-based training. As our results show, LASP with virtual classes can significantly improve the robustness of the prompts learned. We refer to this variant of our method as \textbf{\shortNameZero}.

Note that training with virtual classes does not violate the zero-shot setting~\cite{xian2017zero}\footnote{according to~\cite{xian2017zero} ``Zero-shot learning aims to recognize objects whose \textbf{instances} may not have been seen during training.''}. Moreover, from a practical perspective, if the novel class names are not known during initial training, the model can be simply retrained in a zero-shot manner when they become available.

\section{Experiments}\label{sec:experiments}

Following~\cite{radford2019language, zhou2022conditional}, we mainly evaluated the accuracy of our approach on generalization to novel classes (i.e. zero-shot recognition) for 11 datasets in total. Each dataset is split into two equal partitions with disjoint classes, named \textit{base} and \textit{new}. We trained our model using text-image pairs from the base classes and test on both base and new classes. For other types of experiments including cross-dataset transfer and domain generalization, see Supp. Mat.

\noindent \textbf{Datasets:} We used 11 in total, namely: ImageNet~\cite{deng2009imagenet}, Caltech101~\cite{fei2004learning}, Oxford-Pets~\cite{parkhi2012cats}, Stanford Cars~\cite{krause20133d}, Flowers102~\cite{nilsback2008automated}, Food101~\cite{bossard2014food}, FGVC Aircraft~\cite{maji2013fine}, SUN397~\cite{xiao2010sun}, DTD~\cite{cimpoi2014describing}, EuroSAT~\cite{helber2019eurosat} and UCF-101~\cite{soomro2012ucf101}.

\noindent \textbf{Models:} For all experiments, unless otherwise specified, we used a pretrained CLIP model with a ViT-B/16 image encoder, $M=4$ learnable prompts and 16 samples per class. The number of groups $G$ (when used) is set to 3. In all experiments, we report the average across 3 runs.

\noindent \textbf{Training:} Largely, we followed the training procedure described in CoOp~\cite{zhou2022learning} and CoCoOp~\cite{zhou2022conditional} (i.e. same image augmentation, SGD with initial learning rate of $0.002$ and a cosine annealing scheduler with 1 epoch of warm-up). In Eq.~\ref{eq:loss_combined}, $\alpha_{VL}$ was set to 1 and $\alpha_{T}$ to 20. The number of textual templates $L$ was set to 34. The templates were taken from CoOp and CLIP (see Supp. Mat. for a full list). All training and testing was done on a single NVIDIA V100 GPU (except for Imagenet where 4 GPUs were used). The code was implemented using PyTorch~\cite{paszke2017automatic}.

\noindent \textbf{Methods compared:} We report the performance of LASP and its improved version trained with virtual classes (LASP-V).
For LASP-V, the \textit{class names only} of the novel classes are used during training as virtual classes. We also study the impact of adding other types of virtual classes. The direct baseline that our method is compared with is CoOp~\cite{zhou2022learning}, as we add the proposed components on top of it. Note that both methods have \textit{exactly} the same inference (as our method adds in addition a text-to-text loss during training). We also compare with ProDA~\cite{lu2022prompt}  and CoCoOp~\cite{zhou2022conditional} which conditions the prompts on image features and hence induces significant additional computation during inference.

\subsection{Comparison with state-of-the-art}

\noindent \textbf{Standard setting of~\cite{zhou2022conditional}:} Table~\ref{tab:results_generalization_sota} compares our approach with the current state-of-the-art. We conclude:
\vspace{-0.25cm}
\begin{itemize}
\item
\textbf{Conclusion 1: In terms of harmonic mean, LASP outperforms all methods by large margin.} It outperforms, on average, the second best (ProDA) by $>2\%$. The improvement on specific datasets is even bigger (e.g. $>3\%$ on Flowers102, $>11\%$ on EuroSAT, $>3\%$ on UCF101).
\item
\textbf{Conclusion 2: On the novel classes, LASP outperforms all methods by large margin.} It is the first reported method outperforming CLIP by 0.68\% (but notice that CLIP performs very poorly on the bases classes). It also outperforms ProDA (third best) by $>2.5\%$. Again, compared to ProDA, the improvement on specific datasets is even bigger (e.g. $>5\%$ on Flowers102, $>3\%$ on Food101, $>11\%$ on EuroSAT, $>6\%$ on UCF101). 
\item 
\textbf{Conclusion 3: On new classes, LASP with virtual classes has significant impact  for specific datasets}. These include datasets with informative class names like EuroSAT and DTD where the improvement over LASP is $\sim5.5\%$ and $\sim4.0\%$, respectively.
\end{itemize}

\begin{table}[ht]
    \tabstyle{1pt}
    \centering
    \caption{\textbf{Comparison with the state-of-the-art on 11 datasets}. We provide the results of LASP and LASP trained with virtual classes (LASP-V). $\Delta$ denotes the absolute improvement of our best variant, LASP-V, over the previous best result.}\label{tab:results_generalization_sota}
    \resizebox{\linewidth}{!}{%
    \begin{tabular}{l ccccc|cc|c}
    \toprule
    \multirow{2}{*}{Dataset} &  \multirow{2}{*}{Set} & CLIP & CoOp & CoCoOp & ProDA & LASP & LASP-V & \multirow{2}{*}{$\Delta$} \\
    & & \cite{radford2021learning,zhou2022learning} & \cite{zhou2022learning} & \cite{zhou2022conditional} & \cite{lu2022prompt}  & (Ours) & (Ours) &\\
    \midrule
    \multirow{3}{*}{Average}&Base & 69.34 & 82.69 & 80.47 & 81.56 & 82.70 & \textbf{83.18} & \hgreen{+0.49} \\
    &New & 74.22 & 63.22 & 71.69 & 72.30  & 74.90 & \textbf{76.11} & \hgreen{+1.89} \\
    &H & 71.70 & 71.66 & 75.83 & 76.65  & 78.61 & \textbf{79.48} & \hgreen{+2.83} \\
    \midrule
    \multirow{3}{*}{ImageNet}&Base & 72.43 & \textbf{76.47} & 75.98 & 75.40 & 76.20 & 76.25 & \textbf{\color{carnelian}-0.22}\\
    &New & 68.14 & 67.88 & 70.43 & 70.23  & 70.95 & \textbf{71.17} & \hgreen{+0.74}\\
    &H & 70.22 & 71.92 & 73.10 & 72.72   & 73.48 & \textbf{73.62} & \hgreen{+0.52}\\
    \midrule
    \multirow{3}{*}{Caltech101}&Base & 96.84 & 98.0 & 97.96 & \textbf{98.27}  & 98.10 & 98.17 & \textbf{\color{carnelian}-0.10}\\
    &New & 94.0 & 89.91 & 93.81 & 93.23  & 94.24 & \textbf{94.33} & \hgreen{+0.33} \\
    &H & 95.40 & 93.73 & 95.84 & 95.86  & 96.16 & \textbf{96.21} & \hgreen{+0.35} \\
    \midrule
    \multirow{3}{*}{OxfordPets}&Base & 91.17 & 93.67 & 95.20 & 95.43  & \textbf{95.90} & 95.73 & \hgreen{+0.30}\\
    &New & 97.26 & 95.29 & 97.69 & 97.83  & \textbf{97.93} & 97.87 & \hgreen{+0.04} \\
    &H & 94.12 & 94.47 & 96.43 & 96.62  & \textbf{96.90} & 96.79 & \hgreen{+0.16}  \\
    \midrule
    \multirow{3}{*}{
        \begin{tabular}{l}Stanford\\
       Cars
    \end{tabular}
    }&Base & 63.37 & \textbf{78.12} & 70.49 & 74.70  & 75.17 & 75.23 & {\color{carnelian} \textbf{-2.89}} \\
    &New & \textbf{74.89} & 60.40 & 73.59 & 71.20  & 71.60 & 71.77 & {\color{carnelian} \textbf{-3.12}}\\
    &H & 68.85 & 68.13& 72.01 & 72.91  & 73.34 & \textbf{73.46} & \hgreen{+0.55}\\
    \midrule
    \multirow{3}{*}{Flowers102}&Base & 72.08 & 97.60 & 94.87 & \textbf{97.70}  & 97.0 & 97.17 & {\color{carnelian} \textbf{-0.53}} \\
    &New & \textbf{77.80} & 59.67 & 71.75 & 68.68  & 74.0 & 73.53 & {\color{carnelian} \textbf{-4.27}}\\
    &H & 74.83 & 74.06 & 81.71 & 80.66  & 83.95 & \textbf{83.71} & \hgreen{+2.0}\\ 
    \midrule
    \multirow{3}{*}{Food101}&Base & 90.10 & 88.33 & 90.70 & 90.30  & 91.20 & \textbf{91.20} & \hgreen{+0.50}\\
    &New & 91.22 & 82.26 & 91.29 & 88.57  & 91.70 & \textbf{91.90} & \hgreen{+0.61}\\
    &H & 90.66 & 85.19 & 90.99 & 89.43  & 91.44 & \textbf{91.54} & \hgreen{+0.55} \\ 
    \midrule
    \multirow{3}{*}{
    \begin{tabular}{l}FGVC\\
       Aircraft
    \end{tabular}
    }&Base & 27.19 & \textbf{40.44} & 33.41 & 36.90  & 34.53 &  38.05 & {\color{carnelian} \textbf{-2.39}} \\
    &New & \textbf{36.29} & 22.3 & 23.71 & 34.13 &  30.57 & 33.20 & {\color{carnelian}\textbf{-3.09}}\\
    &H & 31.09 & 28.75 & 27.74 & \textbf{35.46} & 32.43 & \textbf{35.46} & \textbf{\color{gray}0.0} \\ 
    \midrule
    \multirow{3}{*}{SUN397}&Base & 69.36 & 80.6 & 79.74 & 78.67  & 80.70 & \textbf{80.70} & \hgreen{+0.10} \\
    &New & 75.35 & 65.89 & 76.86 & 76.93  & 78.60 & \textbf{79.30} & \hgreen{+2.37}\\
    &H & 72.23 & 72.51 & 78.27 & 77.79  & 79.63 & \textbf{80.0} & \hgreen{+1.73} \\ 
    \midrule
    \multirow{3}{*}{DTD}&Base & 53.24 & 79.44 & 77.01 & 80.67  & 81.4 & \textbf{81.10}  & \hgreen{+1.53} \\
    &New & 59.9 & 41.18 & 56.0 & 56.48 &  58.6  & \textbf{62.57} & \hgreen{+3.10}\\
    &H & 56.37 & 54.24 & 64.85 & 66.44 &  68.14 & \textbf{70.64}  &\hgreen{+4.20} \\ 
    \midrule
    \multirow{3}{*}{EuroSAT}&Base & 56.48 & 92.19 & 87.49 & 83.90 & 94.60 & \textbf{95.0} & \hgreen{+2.81} \\
    &New & 64.05 & 54.74 & 60.04 & 66.0 & 77.78 & \textbf{83.37} & \hgreen{+17.37}\\
    &H & 60.03 & 68.9 & 71.21 & 73.88  & 85.36 & \textbf{88.86} & \hgreen{+14.98}\\ 
    \midrule
    \multirow{3}{*}{UCF101}&Base & 70.53 & 84.69 & 82.33 & 85.23  & 84.77 & \textbf{85.53} & \hgreen{+0.30}\\
    &New & 77.50 & 56.05 & 73.45 & 71.97  & 78.03 & \textbf{78.20} & \hgreen{+0.70} \\
    &H & 73.85 & 67.46 & 77.64 & 78.04 & 81.26 & \textbf{81.70} & \hgreen{+3.66}\\ 

    \bottomrule
    \end{tabular}
    }
\vspace{-0.35cm}
\end{table}

\begin{table}[ht]
    \tabstyle{2pt}
    \centering
    \caption{\textbf{Effect of different LASP components.} Text-to-Text is Eq.~\ref{eq:Ltt}, only. On top of this, we incrementally apply the grouped prompt of Eq.~\ref{eq:Lttg}, and the re-alignment module of~Sec.~\ref{ssec:method_vision_distribution}. Up to this point, this is equiv. to LASP. Finally, we add virtual classes (equiv. to LASP-V). Baseline is CoOp.     }\label{tab:results_generalization_ablation}
        \resizebox{\linewidth}{!}{%
    \begin{tabular}{l cc|cccc}
    \toprule
    \multirow{2}{*}{Dataset} &  \multirow{2}{*}{Set}  & Baseline  & Text-to-Text  & +Grouped & +Align & + Virtual  \\
    & & \cite{zhou2022learning}  & &    & (LASP) & (LASP-V) \\
    \midrule
    \multirow{3}{*}{Average}&Base &  82.69  & 81.26 &  81.87 & 82.70 & \textbf{83.18}  \\
    &New  & 63.22 &  71.54  & 73.48 & 74.90 & \textbf{76.11}  \\
    &H  & 71.66  & 76.09  & 77.44 & 78.61 & \textbf{79.48}  \\
    \midrule
    \multirow{3}{*}{ImageNet}&Base &  \textbf{76.47} & 75.97  & 76.20 & 76.20 & 76.25 \\
    &New  & 67.88 & 70.31  & 70.70 & 70.95 & \textbf{71.17} \\
    &H  & 71.92   & 73.03  &  73.34 &73.48 & \textbf{73.62} \\
    \midrule
    \multirow{3}{*}{Caltech101}&Base  & 98.0  & 97.70  & 97.97 & 98.10 & \textbf{98.17} \\
    &New  & 89.91 &  94.08  & 94.27 & 94.24 & \textbf{94.33}  \\
    &H &  93.73  & 95.85  & 96.08 & 96.16 & \textbf{96.21}  \\
    \midrule
    \multirow{3}{*}{OxfordPets}&Base  & 93.67  & 95.13  & 95.63 & \textbf{95.90} & 95.73 \\
    &New  & 95.29  & 96.23 & 97.87 & \textbf{97.93} & 97.87  \\
    &H  & 94.47  & 95.68 & 96.73 & \textbf{96.90} & 96.79   \\
    \midrule
    \multirow{3}{*}{
        \begin{tabular}{l}Stanford\\
       Cars
    \end{tabular}
    }&Base  & \textbf{78.12}  & 72.46  & 73.50 & 75.17 & 75.23  \\
    &New  & 60.40  & 71.80  & \textbf{72.1} & 71.60 & 71.77 \\
    &H  & 68.13& 72.19 & 72.93 &73.34 & \textbf{73.46} \\
    \midrule
    \multirow{3}{*}{Flowers102}&Base  & \textbf{97.60} & 96.47  & 96.80 & 97.0 & 97.17  \\
    &New &  59.67  & 70.7  & \textbf{74.0} & \textbf{74.0} & 73.53 \\
    &H  & 74.06  & 81.59 & 83.87 & \textbf{83.95} & 83.71 \\ 
    \midrule
    \multirow{3}{*}{Food101}&Base  & 88.33  & 90.30  & 91.0 & \textbf{91.20} & \textbf{91.20} \\
    &New  & 82.26  & 90.73  & 90.87 & 91.70 & \textbf{91.90} \\
    &H  & 85.19  & 90.51  & 90.93 & 91.44 & \textbf{91.54}  \\ 
    \midrule
    \multirow{3}{*}{
    \begin{tabular}{l}FGVC\\
       Aircraft
    \end{tabular}
    }&Base  & \textbf{40.44}  & 32.63  & 33.05 & 34.53 &  38.05  \\
    &New &  22.3  & 30.46  &  31.80 & 30.57 & \textbf{33.20} \\
    &H &  28.75  & 31.57   & 32.41 & 32.43 & \textbf{35.46}  \\ 
    \midrule
    \multirow{3}{*}{SUN397}&Base  & 80.6  & 80.20  & 80.55 & \textbf{80.70} & \textbf{80.70}  \\
    &New  & 65.89  & 75.56  & 77.11 & 78.60 & \textbf{79.30} \\
    &H  & 72.51  & 77.81  & 78.79 & 79.63 & \textbf{80.0}  \\ 
    \midrule
    \multirow{3}{*}{DTD}&Base & 79.44  & 79.13  & 80.5 & \textbf{81.4} & 81.10   \\
    &New &  41.18  & 52.1 &  56.20 & 58.6  & \textbf{62.57} \\
    &H  & 54.24  & 62.82 &  66.19 & 68.14 & \textbf{70.64}  \\ 
    \midrule
    \multirow{3}{*}{EuroSAT}&Base  & 92.19  & 91.23  & 91.90 & 94.60 & \textbf{95.0}  \\
    &New  & 54.74  & 63.16  & 66.37 & 77.78 & \textbf{83.37} \\
    &H  & 68.9  & 74.64  & 77.07 & 85.36 & \textbf{88.86} \\ 
    \midrule
    \multirow{3}{*}{UCF101}&Base &  84.69  & 82.7  & 83.47 &84.77 & \textbf{85.53}\\
    &New  & 56.05  & 71.80  & 77.07 & 78.03 & \textbf{78.20}  \\
    &H  & 67.46  & 76.86  & 80.14 & 81.26 & \textbf{81.70} \\ 

    \bottomrule
    \end{tabular}
}
\end{table}

\noindent \textbf{Generalized zero-shot setting:} The current evaluation protocol used in~\cite{zhou2022conditional} computes the accuracy considering the base and new classes in isolation. A more realistic evaluation protocol should consider the classes across both subsets (i.e. base and novel) jointly. Detailed results for this setting are provided in the Supp. Mat., but, in general, the same conclusions as above hold.

\subsection{Ablation studies}~\label{ssec:ablation}

\noindent \textbf{Effect of different LASP components:}
LASP proposes a number of contributions which are evaluated incrementally. The start point is the proposed Text-to-Text loss of Eq.~\ref{eq:Ltt}. On top of this, we incrementally apply the grouped prompt representation (Eq.~\ref{eq:Lttg}), and then the re-alignment module (Sec.~\ref{ssec:method_vision_distribution}). This gives rise to LASP. Finally, we add virtual classes giving rise to LASP-V. Our baseline is CoOp. From the results of  Table~\ref{tab:results_generalization_ablation}, we conclude:
\vspace{-0.25cm}
\begin{itemize}
\item
\textbf{Conclusion 4: Our idea in its plain form (Text-to-Text loss) outperforms its direct baseline (CoOp) by a large margin}. Specifically, it improves upon CoOp by $\sim4.5\%$ on average, demonstrating its effectiveness. 
\item
\textbf{Conclusion 5: All components are needed to obtain high accuracy.} 
\end{itemize}

\noindent \textbf{Effect of size and content of the textual prompts:} Herein, we study the effect of the size $L$ and the content of the set of the textual prompts used by our method in Eq.~\ref{eq:p_t}. For simplicity, we report results using our Text-to-Text loss (Eq.~\ref{eq:Ltt}), only. The hand-crafted templates are increased to 100 by including the rest of the prompts defined in CLIP~\cite{radford2021learning}, while their number is reduced to 1 by using the following template only: \texttt{a photo of \{\}}. Random templates are produced by sampling grammatically plausible random sentences that contain incoherent words, with length between 5 and 20 words. The class names are inserted at the end of these random templates (for examples, see Supp. Mat.). All variations use the same training scheduler and hyperparameters, except for the case of random templates, where $\alpha_{TT}=5$. 

Table~\ref{tab:ablation_dict_size} shows our results. We importantly note that the accuracy on the base classes remains similar across all settings (not shown in the table). Moreover, we conclude:
\vspace{-0.25cm}
\begin{itemize}
\item
\textbf{Conclusion 6: The exact choice of the templates might not be so significant for the few-shot setting.}  
\item
\textbf{Conclusion 7: For the case of novel classes, both the number and the content of the templates are important  to obtain high accuracy.}
\end{itemize}

\noindent \textbf{Effect of type of loss:} In Table~\ref{tab:loss_impact_ablation}, we vary the choice of loss in LASP, \ie we replace the Cross-Entropy (CE) with an $L_2$ and $L_1$ loss. Again, for simplicity, we report results using our Text-to-Text loss (Eq.~\ref{eq:Ltt}), only. 
\vspace{-0.25cm}
\begin{itemize}
\item
\textbf{Conclusion 8: The proposed CE loss based formulation outperforms other losses for LASP.} 
\end{itemize}

\begin{table*}[!ht]
    \tabstyle{4pt}
    \caption{\textbf{Effect of dictionary size and content on new classes.} Accuracy on the base classes remains similar across all settings, hence it is omitted. 34 templates were used for the paper's main results. For simplicity, we report results using our Text-to-Text loss (Eq.~\ref{eq:Ltt}), only. Text-to-Text~(R) denotes models trained using randomly constructed templates.}
    \label{tab:ablation_dict_size}
    \begin{subtable}[t]{.32\textwidth}
    \centering
    \caption{DTD.}
    \begin{tabular}{l ccc}
    \toprule
    \#Templates & 1 & 34 & 100 \\
    \midrule
    Text-to-Text~(R) & 49.02 & 51.63 & 52.64  \\
    Text-to-Text & \textbf{50.73} & \textbf{52.10} & \textbf{56.53}  \\
    \bottomrule
    \end{tabular}
    \end{subtable}
    \hfill
    \begin{subtable}[t]{.32\textwidth}
    \centering
    \caption{EuroSAT.}
    \begin{tabular}{l ccc}
    \toprule
    \#Templates & 1 & 34 & 100 \\
    \midrule
    Text-to-Text~(R) & 55.01 & 59.9 & 62.1  \\
    Text-to-Text & \textbf{56.97} & \textbf{63.16} & \textbf{65.13} \\
    \bottomrule
    \end{tabular}
    \end{subtable}
    \hfill
    \begin{subtable}[t]{.32\textwidth}
    \centering
    \caption{UCF101.}
    \begin{tabular}{l ccc}
    \toprule
    \#Templates & 1 & 34 & 100 \\
    \midrule
    Text-to-Text~(R) & 67.5 & 68.6 & 70.03 \\
    Text-to-Text & \textbf{71.36} & \textbf{71.80} & \textbf{72.77} \\
    \bottomrule
    \end{tabular}
    \end{subtable}
\end{table*}

\begin{table*}[!ht]
    \tabstyle{4pt}
    \caption{\textbf{Effect of out-domain distractors.} w/o distractors are the results on the generalized zero-shot setting.}
    \label{tab:results_with_distractors}
    \begin{subtable}[t]{.49\textwidth}
    \centering
    \caption{EuroSAT.}
        \begin{tabular}{l| ccc| ccc}
        \toprule
        Method & \multicolumn{3}{c}{w/o distractors} &
          \multicolumn{3}{c}{with distractors} \\
          \cmidrule{2-7}
        & Base & New & H & Base & New & H \\
        \midrule
        \shortName & 86.25 & 64.63 & 73.89 & 86.0 & 55.80 & 67.68 \\
        \shortNameZero & \textbf{90.0} & \textbf{65.73} & \textbf{75.97} & \textbf{90.8} & \textbf{59.87} & \textbf{72.16}\\
        \bottomrule
        \end{tabular}
    \end{subtable}
     \vspace{1em}
    \hfill
    \begin{subtable}[t]{.49\textwidth}
    \centering
    \caption{Food101.}
    \begin{tabular}{l| ccc| ccc}
    \toprule
    Method & \multicolumn{3}{c}{w/o distractors} &
      \multicolumn{3}{c}{with distractors} \\
      \cmidrule{2-7}
    & Base & New & H & Base & New & H \\
    \midrule
    \shortName & \textbf{87.17} & 87.53 & 87.34 & \textbf{87.01} & 86.90 & 86.95\\
    \shortNameZero & \textbf{87.17} & \textbf{87.63} & \textbf{87.39} & 86.99 & \textbf{87.10} & \textbf{87.04}\\
    \bottomrule
    \end{tabular}
    \end{subtable}
    \hfill
    \begin{subtable}[t]{.49\textwidth}
    \centering
    \caption{Flowers102.}
    \begin{tabular}{l| ccc| ccc}
    \toprule
    Method & \multicolumn{3}{c}{w/o distractors} &
      \multicolumn{3}{c}{with distractors} \\
      \cmidrule{2-7}
    & Base & New & H & Base & New & H \\
    \midrule
    \shortName & 90.97 & 67.8 & 77.69 & 90.0 & 67.1 & 76.68\\
    \shortNameZero & \textbf{93.20} & \textbf{69.93} & \textbf{79.9} & \textbf{92.05} & \textbf{69.08} & \textbf{78.92}\\
    \bottomrule
    \end{tabular}
    \end{subtable}
    \hfill
    \begin{subtable}[t]{.49\textwidth}
    \centering
    \caption{OxfordPets.}
    \begin{tabular}{l| ccc| ccc}
    \toprule
    Method & \multicolumn{3}{c}{w/o distractors} &
      \multicolumn{3}{c}{with distractors} \\
      \cmidrule{2-7}
    & Base & New & H & Base & New & H \\
    \midrule
    \shortName & \textbf{92.53} & \textbf{94.20} & \textbf{91.52} & 91.53 & 92.60 & 92.06\\
    \shortNameZero & 92.25 & 93.97 & 93.10 & \textbf{92.23} & \textbf{93.17} & \textbf{92.69}\\
    \bottomrule
    \end{tabular}
    \end{subtable}
    
\end{table*}

\begin{table*}[!ht]
    \tabstyle{4pt}
    \caption{\textbf{Effect of in-domain distractors.} w/o distractors are the results on the generalized zero-shot setting evaluation.}
    \label{tab:results_with_sim_distractors}
    \begin{subtable}[t]{.49\textwidth}
    \centering
    \caption{Food101.}
    \begin{tabular}{l| ccc| ccc}
    \toprule
    Method & \multicolumn{3}{c}{w/o distractors} &
      \multicolumn{3}{c}{with distractors} \\
      \cmidrule{2-7}
    & Base & New & H & Base & New & H \\
    \midrule
    \shortName & \textbf{87.17} & 87.53 & 87.34 & 82.70 & 83.47 & 83.08\\
    \shortNameZero & \textbf{87.17} & \textbf{87.63} & \textbf{87.39} & \textbf{83.11} & \textbf{83.95} & \textbf{83.52}\\
    \bottomrule
    \end{tabular}
    \end{subtable}
     \vspace{1em}
    \hfill
    \begin{subtable}[t]{.49\textwidth}
    \centering
    \caption{Flowers102.}
    \begin{tabular}{l| ccc| ccc}
    \toprule
    Method & \multicolumn{3}{c}{w/o distractors} &
      \multicolumn{3}{c}{with distractors} \\
      \cmidrule{2-7}
    & Base & New & H & Base & New & H \\
    \midrule
    \shortName & 90.97 & 67.8 & 77.69 & 80.16 & 62.50 & 70.23\\
    \shortNameZero & \textbf{93.20} & \textbf{69.93} & \textbf{79.9} & \textbf{83.95} & \textbf{65.31} & \textbf{73.47}\\
    \bottomrule
    \end{tabular}
    \end{subtable}
    \vspace*{-0.45cm}
\end{table*}

\noindent \textbf{Effect of out-domain distractors:} Motivated by the recent work of~\cite{ren2022rethinking} suggesting that CLIP's performance drops as the number of classes used for testing increases, we introduce a new evaluation setting: Firstly, we select 4 test datasets with clear disjoint domains: EuroSAT (10 satellite terrain types), Food101 (101 food names), Flowers102 (102 flower names) and OxfordPets (37 dog and cat breed names). At test time, we define the classifier across the union of classes across all 4 datasets (250 classes in total). Note that  \shortNameZero~ is the only method that benefits from knowledge of this expanded vocabulary during training. From Table~\ref{tab:results_with_distractors}, we can conclude:
\vspace{-0.25cm}
\begin{itemize}
\item
\textbf{Conclusion 9: The models are somewhat robust to out-of-domain distractors.} Specifically, the drop in accuracy is moderate (typically 1-2\%). The exception is EuroSAT where the number of classes increases $25\times$. Importantly, LASP-V manages to largely recover the lost accuracy.
\end{itemize}

\noindent \textbf{Effect of in-domain distractors:} Expanding on the idea from the previous section, herein, we propose to test the performance of the current soft prompting methods with in-domain distractors. Unlike the case of out-of-domain distractors, the in-domain distractors are selected such that they are closely related to the current dataset/classes being part of the same super-category. We performed experiments on two datasets: Food101 and Flowers102. For Flowers102, we added 65 new class names while, for Food101, 53 new classes. Note again that, except for LASP-V, the classes are only used at test time as distractors expanding the C-way classifier by 65 and 53, respectively. The list of added classes can be found in the Supp. Mat. From the results of Table~\ref{tab:results_with_sim_distractors}, we conclude: 
\vspace{-0.15cm}
\begin{itemize}
\item
\textbf{Conclusion 10: In-domain distractors significantly increase the problem difficulty.} Specifically, the drop in accuracy is large (4-7\%). LASP-V manages to recover part of the lost accuracy.
\end{itemize}

\begin{table}[ht]
    \centering
        \caption{\textbf{Effect of type of loss.} For simplicity, we report results using our Text-to-Text loss (Eq.~\ref{eq:Ltt}), only.}
    \label{tab:loss_impact_ablation}
    \begin{tabular}{c|cccc}
    \toprule
       Set  & CE & $L_1$ & $L_2$  \\
       \midrule
       Base & 81.26 & \textbf{81.50} & 81.47  \\
       New & \textbf{71.54} & 66.01 & 65.80 \\
       H  & \textbf{76.09} & 73.54 & 72.80\\
    \bottomrule
    \end{tabular}
\vspace{-0.2cm}
\end{table}

\section{Conclusions}

In this paper, we introduced LASP - a language aware soft prompting method for V\&L adaptation that is shown to outperform prior work by large margin. Specifically, we made the following contributions: \textit{Firstly}, we introduced a novel text-to-text loss that largely alleviates the problem of base-class overfitting. \textit{Secondly}, we proposed a \textit{grouped} language-aware prompting for learning more specialized and stronger prompt representations. \textit{Thirdly}, we identified a visual-language misalignment within LASP and propose a re-calibration mechanism to address it.
\textit{Fourthly}, we showed that our approach, unlike prior work, is amenable to, including during training, \textit{virtual classes}, i.e. class names for which no visual samples are available, significantly increasing the robustness of the learned prompts. We hope that LASP/LASP-V will serve as a strong baseline for future works in the area of few-shot adaptation for V\&L models.

\clearpage

{\small
\bibliographystyle{ieee_fullname}
\bibliography{egbib}

\begin{thebibliography}{10}\itemsep=-1pt

\bibitem{alayrac2022flamingo}
Jean-Baptiste Alayrac, Jeff Donahue, Pauline Luc, Antoine Miech, Iain Barr,
  Yana Hasson, Karel Lenc, Arthur Mensch, Katie Millican, Malcolm Reynolds,
  et~al.
\newblock Flamingo: a visual language model for few-shot learning.
\newblock {\em arXiv preprint arXiv:2204.14198}, 2022.

\bibitem{ba2016layer}
Jimmy~Lei Ba, Jamie~Ryan Kiros, and Geoffrey~E Hinton.
\newblock Layer normalization.
\newblock {\em arXiv preprint arXiv:1607.06450}, 2016.

\bibitem{bossard2014food}
Lukas Bossard, Matthieu Guillaumin, and Luc~Van Gool.
\newblock Food-101--mining discriminative components with random forests.
\newblock In {\em European conference on computer vision}, pages 446--461.
  Springer, 2014.

\bibitem{brown2020language}
Tom Brown, Benjamin Mann, Nick Ryder, Melanie Subbiah, Jared~D Kaplan, Prafulla
  Dhariwal, Arvind Neelakantan, Pranav Shyam, Girish Sastry, Amanda Askell,
  et~al.
\newblock Language models are few-shot learners.
\newblock {\em Advances in neural information processing systems},
  33:1877--1901, 2020.

\bibitem{cimpoi2014describing}
Mircea Cimpoi, Subhransu Maji, Iasonas Kokkinos, Sammy Mohamed, and Andrea
  Vedaldi.
\newblock Describing textures in the wild.
\newblock In {\em Proceedings of the IEEE conference on computer vision and
  pattern recognition}, pages 3606--3613, 2014.

\bibitem{deng2009imagenet}
Jia Deng, Wei Dong, Richard Socher, Li-Jia Li, Kai Li, and Li Fei-Fei.
\newblock Imagenet: A large-scale hierarchical image database.
\newblock In {\em 2009 IEEE conference on computer vision and pattern
  recognition}, pages 248--255. Ieee, 2009.

\bibitem{devlin2018bert}
Jacob Devlin, Ming-Wei Chang, Kenton Lee, and Kristina Toutanova.
\newblock Bert: Pre-training of deep bidirectional transformers for language
  understanding.
\newblock {\em arXiv preprint arXiv:1810.04805}, 2018.

\bibitem{dosovitskiy2020image}
Alexey Dosovitskiy, Lucas Beyer, Alexander Kolesnikov, Dirk Weissenborn,
  Xiaohua Zhai, Thomas Unterthiner, Mostafa Dehghani, Matthias Minderer, Georg
  Heigold, Sylvain Gelly, et~al.
\newblock An image is worth 16x16 words: Transformers for image recognition at
  scale.
\newblock {\em arXiv preprint arXiv:2010.11929}, 2020.

\bibitem{fei2004learning}
Li Fei-Fei, Rob Fergus, and Pietro Perona.
\newblock Learning generative visual models from few training examples: An
  incremental bayesian approach tested on 101 object categories.
\newblock In {\em 2004 conference on computer vision and pattern recognition
  workshop}, pages 178--178. IEEE, 2004.

\bibitem{helber2019eurosat}
Patrick Helber, Benjamin Bischke, Andreas Dengel, and Damian Borth.
\newblock Eurosat: A novel dataset and deep learning benchmark for land use and
  land cover classification.
\newblock {\em IEEE Journal of Selected Topics in Applied Earth Observations
  and Remote Sensing}, 12(7):2217--2226, 2019.

\bibitem{hinton2015distilling}
Geoffrey Hinton, Oriol Vinyals, Jeff Dean, et~al.
\newblock Distilling the knowledge in a neural network.
\newblock {\em arXiv preprint arXiv:1503.02531}, 2(7), 2015.

\bibitem{unsup_prompt22}
Tony Huang, Jack Chu, and Fangyun Wei.
\newblock Unsupervised prompt learning for vision-language models.
\newblock {\em arXiv preprint arXiv:2204.03649}, 2022.

\bibitem{jia2021scaling}
Chao Jia, Yinfei Yang, Ye Xia, Yi-Ting Chen, Zarana Parekh, Hieu Pham, Quoc Le,
  Yun-Hsuan Sung, Zhen Li, and Tom Duerig.
\newblock Scaling up visual and vision-language representation learning with
  noisy text supervision.
\newblock In {\em International Conference on Machine Learning}, pages
  4904--4916. PMLR, 2021.

\bibitem{krause20133d}
Jonathan Krause, Michael Stark, Jia Deng, and Li Fei-Fei.
\newblock 3d object representations for fine-grained categorization.
\newblock In {\em Proceedings of the IEEE international conference on computer
  vision workshops}, pages 554--561, 2013.

\bibitem{krizhevsky2017imagenet}
Alex Krizhevsky, Ilya Sutskever, and Geoffrey~E Hinton.
\newblock Imagenet classification with deep convolutional neural networks.
\newblock {\em Communications of the ACM}, 60(6):84--90, 2017.

\bibitem{lester2021power}
Brian Lester, Rami Al-Rfou, and Noah Constant.
\newblock The power of scale for parameter-efficient prompt tuning.
\newblock {\em arXiv preprint arXiv:2104.08691}, 2021.

\bibitem{li2021prefix}
Xiang~Lisa Li and Percy Liang.
\newblock Prefix-tuning: Optimizing continuous prompts for generation.
\newblock {\em arXiv preprint arXiv:2101.00190}, 2021.

\bibitem{li2021supervision}
Yangguang Li, Feng Liang, Lichen Zhao, Yufeng Cui, Wanli Ouyang, Jing Shao,
  Fengwei Yu, and Junjie Yan.
\newblock Supervision exists everywhere: A data efficient contrastive
  language-image pre-training paradigm.
\newblock {\em arXiv preprint arXiv:2110.05208}, 2021.

\bibitem{lu2022prompt}
Yuning Lu, Jianzhuang Liu, Yonggang Zhang, Yajing Liu, and Xinmei Tian.
\newblock Prompt distribution learning.
\newblock In {\em IEEE Conference on Computer Vision and Pattern Recognition},
  2022.

\bibitem{maji2013fine}
Subhransu Maji, Esa Rahtu, Juho Kannala, Matthew Blaschko, and Andrea Vedaldi.
\newblock Fine-grained visual classification of aircraft.
\newblock {\em arXiv preprint arXiv:1306.5151}, 2013.

\bibitem{nilsback2008automated}
Maria-Elena Nilsback and Andrew Zisserman.
\newblock Automated flower classification over a large number of classes.
\newblock In {\em 2008 Sixth Indian Conference on Computer Vision, Graphics \&
  Image Processing}, pages 722--729. IEEE, 2008.

\bibitem{parkhi2012cats}
Omkar~M Parkhi, Andrea Vedaldi, Andrew Zisserman, and CV Jawahar.
\newblock Cats and dogs.
\newblock In {\em 2012 IEEE conference on computer vision and pattern
  recognition}, pages 3498--3505. IEEE, 2012.

\bibitem{paszke2017automatic}
Adam Paszke, Sam Gross, Soumith Chintala, Gregory Chanan, Edward Yang, Zachary
  DeVito, Zeming Lin, Alban Desmaison, Luca Antiga, and Adam Lerer.
\newblock Automatic differentiation in pytorch.
\newblock 2017.

\bibitem{radford2021learning}
Alec Radford, Jong~Wook Kim, Chris Hallacy, Aditya Ramesh, Gabriel Goh,
  Sandhini Agarwal, Girish Sastry, Amanda Askell, Pamela Mishkin, Jack Clark,
  et~al.
\newblock Learning transferable visual models from natural language
  supervision.
\newblock In {\em International Conference on Machine Learning}, pages
  8748--8763. PMLR, 2021.

\bibitem{radford2019language}
Alec Radford, Jeffrey Wu, Rewon Child, David Luan, Dario Amodei, Ilya
  Sutskever, et~al.
\newblock Language models are unsupervised multitask learners.
\newblock {\em OpenAI blog}, 1(8):9, 2019.

\bibitem{ren2022rethinking}
Shuhuai Ren, Lei Li, Xuancheng Ren, Guangxiang Zhao, and Xu Sun.
\newblock Rethinking the openness of clip.
\newblock {\em arXiv preprint arXiv:2206.01986}, 2022.

\bibitem{schick2020exploiting}
Timo Schick and Hinrich Sch{\"u}tze.
\newblock Exploiting cloze questions for few shot text classification and
  natural language inference.
\newblock {\em arXiv preprint arXiv:2001.07676}, 2020.

\bibitem{schick2020s}
Timo Schick and Hinrich Sch{\"u}tze.
\newblock It's not just size that matters: Small language models are also
  few-shot learners.
\newblock {\em arXiv preprint arXiv:2009.07118}, 2020.

\bibitem{soomro2012ucf101}
Khurram Soomro, Amir~Roshan Zamir, and Mubarak Shah.
\newblock Ucf101: A dataset of 101 human actions classes from videos in the
  wild.
\newblock {\em arXiv preprint arXiv:1212.0402}, 2012.

\bibitem{vaswani2017attention}
Ashish Vaswani, Noam Shazeer, Niki Parmar, Jakob Uszkoreit, Llion Jones,
  Aidan~N Gomez, {\L}ukasz Kaiser, and Illia Polosukhin.
\newblock Attention is all you need.
\newblock {\em Advances in neural information processing systems}, 30, 2017.

\bibitem{xian2017zero}
Yongqin Xian, Bernt Schiele, and Zeynep Akata.
\newblock Zero-shot learning-the good, the bad and the ugly.
\newblock In {\em Proceedings of the IEEE conference on computer vision and
  pattern recognition}, pages 4582--4591, 2017.

\bibitem{xiao2010sun}
Jianxiong Xiao, James Hays, Krista~A Ehinger, Aude Oliva, and Antonio Torralba.
\newblock Sun database: Large-scale scene recognition from abbey to zoo.
\newblock In {\em 2010 IEEE computer society conference on computer vision and
  pattern recognition}, pages 3485--3492. IEEE, 2010.

\bibitem{yao2021filip}
Lewei Yao, Runhui Huang, Lu Hou, Guansong Lu, Minzhe Niu, Hang Xu, Xiaodan
  Liang, Zhenguo Li, Xin Jiang, and Chunjing Xu.
\newblock Filip: Fine-grained interactive language-image pre-training.
\newblock {\em arXiv preprint arXiv:2111.07783}, 2021.

\bibitem{yu2022coca}
Jiahui Yu, Zirui Wang, Vijay Vasudevan, Legg Yeung, Mojtaba Seyedhosseini, and
  Yonghui Wu.
\newblock Coca: Contrastive captioners are image-text foundation models.
\newblock {\em arXiv preprint arXiv:2205.01917}, 2022.

\bibitem{zhou2022conditional}
Kaiyang Zhou, Jingkang Yang, Chen~Change Loy, and Ziwei Liu.
\newblock Conditional prompt learning for vision-language models.
\newblock In {\em Proceedings of the IEEE/CVF Conference on Computer Vision and
  Pattern Recognition}, pages 16816--16825, 2022.

\bibitem{zhou2022learning}
Kaiyang Zhou, Jingkang Yang, Chen~Change Loy, and Ziwei Liu.
\newblock Learning to prompt for vision-language models.
\newblock {\em International Journal of Computer Vision}, 130(9):2337--2348,
  2022.

\bibitem{zhu2022prompt}
Beier Zhu, Yulei Niu, Yucheng Han, Yue Wu, and Hanwang Zhang.
\newblock Prompt-aligned gradient for prompt tuning.
\newblock {\em arXiv preprint arXiv:2205.14865}, 2022.

\end{thebibliography}
}

\end{document}